\documentclass[conference]{IEEEtran}
\IEEEoverridecommandlockouts

\usepackage{cite}
\usepackage{amsmath,amssymb,amsfonts}
\usepackage{algorithmic}
\usepackage{url}
\usepackage{graphicx} 
\usepackage{textcomp}
\usepackage{xcolor}
\usepackage{float}
\usepackage{tabularx}
\usepackage{ragged2e}
\def\BibTeX{{\rm B\kern-.05em{\sc i\kern-.025em b}\kern-.08em
    T\kern-.1667em\lower.7ex\hbox{E}\kern-.125emX}}
\begin{document}

\title{Scalable Back-End for an AI-Based Diabetes Prediction Application
}

\author{%
\parbox{0.3\textwidth}{\centering
Henry Anand Septian Radityo\\[0.3ex]
{\small School of Electrical Engineering and Informatics\\
Bandung Institute of Technology\\
Bandung, Indonesia\\
13521004@std.stei.itb.ac.id}
}%
\hfill
\parbox{0.3\textwidth}{\centering
Bernardus Willson\\[0.3ex]
{\small School of Electrical Engineering and Informatics\\
Bandung Institute of Technology\\
Bandung, Indonesia\\
13521021@std.stei.itb.ac.id}
}%
\hfill
\parbox{0.3\textwidth}{\centering
Raynard Tanadi\\[0.3ex]
{\small School of Electrical Engineering and Informatics\\
Bandung Institute of Technology\\
Bandung, Indonesia\\
13521143@std.stei.itb.ac.id}
}\\[2ex]
\hfill
\parbox{0.48\textwidth}{\centering
Latifa Dwiyanti\\[0.3ex]
{\small School of Electrical Engineering and Informatics\\
Bandung Institute of Technology\\
Bandung, Indonesia\\
latifa@informatika.org}
}
\parbox{0.48\textwidth}{\centering
Saiful Akbar\\[0.3ex]
{\small School of Electrical Engineering and Informatics\\
Bandung Institute of Technology\\
Bandung, Indonesia\\
saiful@informatika.org}
}
}

\maketitle
\IEEEoverridecommandlockouts
\IEEEpubid{\makebox[\columnwidth][l]{979-8-3315-7578-6/25/\$31.00~\copyright~2025 IEEE} \hspace{\columnsep}\makebox[\columnwidth]{ }}
\IEEEpubidadjcol

\begin{abstract}
The rising global prevalence of diabetes necessitates early detection to prevent severe complications. While AI-powered prediction applications offer a promising solution, they require a responsive and scalable back-end architecture to serve a large user base effectively. This paper details the development and evaluation of a scalable back-end system designed for a mobile diabetes prediction application. The primary objective was to maintain a failure rate below 5\% and an average latency of under 1000 ms. The architecture leverages horizontal scaling, database sharding, and asynchronous communication via a message queue. Performance evaluation showed that 83\% of the system's features (20 out of 24) met the specified performance targets. Key functionalities such as user profile management, activity tracking, and read-intensive prediction operations successfully achieved the desired performance. The system demonstrated the ability to handle up to 10,000 concurrent users without issues, validating its scalability. The implementation of asynchronous communication using RabbitMQ proved crucial in minimizing the error rate for computationally intensive prediction requests, ensuring system reliability by queuing requests and preventing data loss under heavy load.
\end{abstract}

\begin{IEEEkeywords}
Scalability, Back-End Architecture, Diabetes Prediction, Asynchronous Communication, Microservices, Performance Testing.
\end{IEEEkeywords}

\section{Introduction}

Diabetes mellitus is a major 21st-century chronic disease with a rising global prevalence. The World Health Organization (WHO) identified it as a leading cause of death in 2019, emphasizing the need for early detection \cite{b1}. In Indonesia, about 11.3\% of adults (20.4 million people) were affected in 2024 \cite{b31}. Advances in artificial intelligence (AI) and machine learning have enabled mobile applications that predict diabetes risk from user data such as diet and activity patterns.
However, their effectiveness relies on a robust back-end infrastructure capable of processing large datasets and executing complex models efficiently. Without scalable design, system performance can degrade under heavy loads, leading to poor user experience and reduced trust \cite{b2}.

A significant challenge in modern AI systems is the ``black box'' problem, where complex models provide highly accurate predictions but offer no insight into their decision-making process. This is particularly problematic in critical domains like healthcare. To address this, this project incorporates Explainable AI (XAI), a set of methods that make AI decisions understandable to humans \cite{b3}. By providing explanations for its risk predictions, the application can build user trust and provide actionable insights, which is a key requirement for clinical and personal health tools. However, generating these explanations adds computational overhead, further emphasizing the need for a highly performant and scalable architecture.

This research addresses the technical challenges of building a high-performance back-end that is both scalable and explainable. The core problems are how to design and implement a back-end architecture that effectively integrates an AI model for diabetes prediction with optimal performance, and how to ensure that the system can accommodate substantial user growth. Scalability, in this context, refers to the ability of a system to handle an increasing workload while maintaining acceptable performance levels, and it is not an afterthought but a fundamental property that determines the long-term viability of software. As noted in the literature, scalability testing has two primary purposes (a) to identify the limits of the system, and (b) to verify that the system performs reliably up to those limits~\cite{saha2018practical}. Following this methodology, performance testing was conducted to empirically determine the maximum number of concurrent users the system could sustain without degradation. The results showed that the system reliably handled up to 10,000 concurrent users, thereby establishing this figure as the tested scalability threshold rather than a predefined design target.

This project aimed to design, implement, and evaluate a scalable back-end system for an AI-powered mobile application for early diabetes risk detection. Success was defined by two metrics, a failure rate below 5\% and an average response latency under 1000 milliseconds~\cite{b4}. Performance testing determined the scalability threshold at 10,000 concurrent users, representing the maximum sustainable load before significant degradation rather than a predefined design target.  

This paper presents the final system architecture, the technologies employed, the database design, and the detailed results of a comprehensive performance evaluation.
\section{Literature Study}
\subsection{Scalability}

Scalability refers to a system’s ability to handle changing workloads while maintaining performance and meeting service-level requirements like latency and availability. Xu~\cite{xu2022systemdesign} highlights two main approaches, vertical scaling, which adds resources to a single node but is limited by hardware capacity, and horizontal scaling, which distributes load across multiple nodes and is widely used in modern distributed systems for its flexibility and fault tolerance. Supporting techniques such as load balancing, caching, database sharding, and asynchronous communication further enhance scalability by reducing bottlenecks, improving responsiveness, and sustaining system efficiency as demand increases.

\subsection{Sharding}

Sharding is a horizontal partitioning method that distributes data across multiple servers (shards) based on criteria such as key values or ranges~\cite{ozsu2011distributeddb}. Unlike replication, which duplicates entire datasets, each shard stores only a subset of the data, enabling the system to scale beyond the limits of a single server. This makes sharding essential when vertical scaling becomes insufficient or too costly, as horizontal scalability can be achieved by adding servers. Strategies include range-based, hash-based, and directory-based sharding~\cite{chang2008bigtable}. Despite challenges like data rebalancing, multi-shard queries, and consistency management, sharding remains a key solution for large-scale systems such as social media, e-commerce, and analytics platforms handling petabytes of data and millions of users.  

\subsection{Performance Testing}

Performance testing evaluates an application’s responsiveness, scalability, stability, and resource usage under varying workloads~\cite{meier2007performance}. In back-end systems, it ensures user loads are handled while maintaining acceptable response times and throughput, using methods such as load, stress, volume, and spike testing, with key metrics including response time, throughput, concurrent users, and resource utilization~\cite{jiang2015survey}. In this study, testing was performed with \textit{k6}, a load testing tool by Grafana Labs~\cite{grafanalabs2023k6}, which uses JavaScript-based scripting, integrates with CI/CD pipelines, supports protocols like HTTP/1.1, HTTP/2, WebSockets, and gRPC, and can simulate tens of thousands of virtual users with low overhead. Its integration with platforms like Prometheus and Grafana, along with its ability to reproduce workloads and model realistic user behavior, made it particularly effective for analyzing system performance under different load conditions.

\subsection{Response Time or Latency Rules}

Response time refers to the total time required by a system to respond to a user request. According to Nielsen's research on response time limits, there are several key thresholds for maintaining a good user experience~\cite{b4}:  
\begin{enumerate}
    \item 0.1 seconds (100 ms) – the limit at which users perceive the system as responding instantly.
    \item 1.0 second (1000 ms) – the limit at which users notice a delay, but it does not interrupt their flow of thought.
    \item 10 seconds – the maximum limit to retain a user’s attention before they are likely to abandon the task.
\end{enumerate}

For implementations where response time approaches the higher thresholds, feedback mechanisms (e.g., progress indicators or loading animations) are recommended to prevent users from perceiving the interface as static or unresponsive.  

\subsection{Error Rate}

Error rate measures the percentage of failed requests compared to the total processed in a system, serving as a key indicator of reliability~\cite{kim2016architecture}. The acceptable threshold depends on the domain, mission-critical areas like banking or healthcare require rates below 0.1\%, while e-commerce and consumer apps can tolerate around 1–5\%. In less critical APIs, even up to 10\% may be acceptable if user experience isn’t heavily impacted. These thresholds highlight the balance between reliability, infrastructure complexity, and cost.

\section{Diabetify Application Overview}

Diabetify is a mobile application designed to support early detection of diabetes risk using artificial intelligence. The main feature of the application is its ability to predict the likelihood of diabetes based on user profile information and lifestyle activities. The profile information includes attributes such as age, body mass index (BMI), cholesterol level, hypertension status, history of macrosomic baby, family history of diabetes, and years of smoking. In addition, the application collects activity data such as smoking habits and exercise frequency to provide a more comprehensive prediction.  

From the user perspective, the application provides a simple interface where users can input their personal and lifestyle data, and then receive a risk prediction result. From the back-end perspective, the critical challenge lies in ensuring that the prediction process remains reliable, accurate, and scalable as the number of users grows. This requires a well-designed back-end architecture to handle user requests efficiently, store and process data securely, and deliver prediction results with minimal latency.

\section{System Architecture and Design}

The architecture was developed iteratively through three experiments, each addressing specific bottlenecks.  
(1) The first design used a load balancer, an ML service with gRPC, and a single database, with the objective of establishing a baseline. Testing showed the single database became a major bottleneck.  
(2) The second design introduced database sharding (dual-database setup) to distribute load, which improved scalability but introduced complexity in managing data consistency.  
(3) The third design added asynchronous processing with RabbitMQ and a cache on top of the sharded databases, which successfully reduced latency, improved reliability, and delivered the best scalability among all versions.
\subsection{Technology Stack}
The choice of technology was critical for achieving the project's performance goals.
\begin{enumerate}
    \item Go (Golang): Used for developing the back-end services due to its efficient concurrency model and low-latency garbage collector, allowing thousands of simultaneous connections with minimal overhead. Benchmarks show it achieves significantly higher throughput compared to Node.js and PHP~\cite{donovan2015go}.
    
    \item PostgreSQL with PGBouncer: PostgreSQL was selected as the relational database for its reliability, strong data integrity, and support for complex data types~\cite{drake2020postgresql}. To address its process-per-connection model, PGBouncer was added as a lightweight connection pooler, enabling hundreds of client connections to be managed efficiently with fewer database connections~\cite{pgbouncer2023docs}.
    
    \item RabbitMQ: Implemented as a message broker using AMQP, RabbitMQ facilitated asynchronous communication, ensured message persistence, and supported flexible routing. This made the system more resilient and prevented long-running tasks from blocking application threads~\cite{videla2012rabbitmq}.
    
    \item Nginx: Deployed as a reverse proxy and load balancer, Nginx leveraged its event-driven architecture to handle thousands of concurrent requests with low resource usage, making it ideal for traffic distribution across service instances~\cite{tanenbaum2016distributed}.
    
    \item Redis: Used as a fast, in-memory cache to store results for “what-if” prediction scenarios, Redis reduced expensive recomputations and offloaded transient queries from the primary database~\cite{redis2024about}.
\end{enumerate}

\subsection{Evaluation Plan}

The evaluation consisted of two parts:  

\begin{enumerate}
    \item Performance Testing: Each of the 24 main API endpoints was subjected to incremental load testing, starting from 1,000 virtual users and increasing until performance degraded significantly. The primary metrics measured were average response time (latency), requests per second (throughput), and failure rate.
    \item Functional Testing: Unit tests were implemented for all controller functions using the Testify library in Go to ensure correctness of business logic in isolation.
\end{enumerate}

\subsection{Architectural Evolution}
The first architectural design divided the system into three main components: a back-end service, a machine learning service, and a centralized database. As shown in Figure~1, a load balancer distributed HTTP requests to multiple back-end instances, which then communicated with the machine learning service via gRPC to reduce latency.  

The main advantages were:  
\begin{enumerate}
    \item Service scalability: the load balancer allowed horizontal scaling of the back-end by adding more instances.  
    \item Separation of concern: business logic was handled in the back-end, while predictions were managed by the machine learning node, simplifying development and maintenance.  
\end{enumerate}
\begin{figure}[H]
\centerline{\includegraphics[width=\columnwidth]{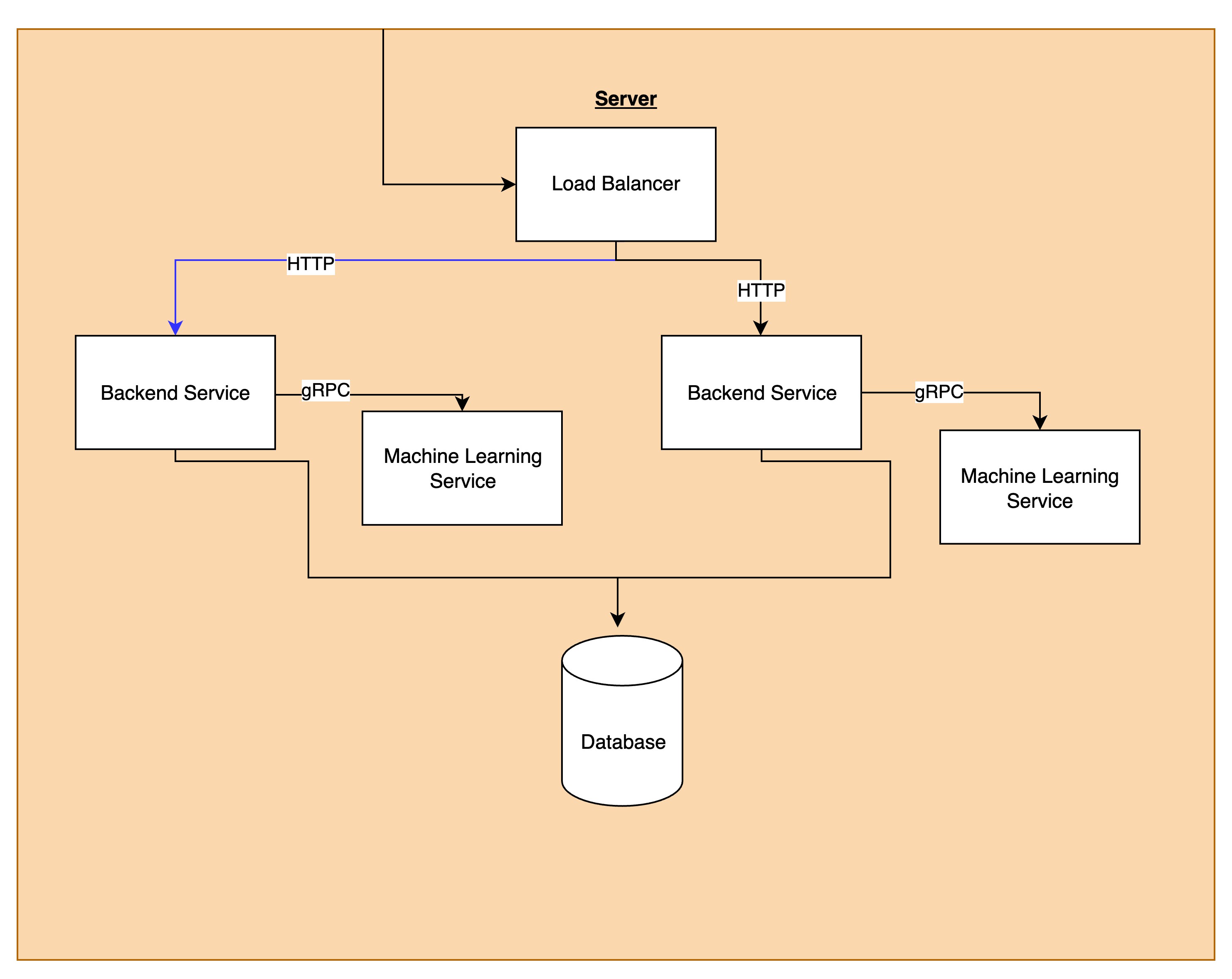}}
\caption{Architecture of Solution 1}
\label{fig_arch}
\end{figure}

Performance testing showed that relying on a single database instance created a major bottleneck, as all read and write operations converged on one server, causing severe performance degradation under high concurrency. A connection pooler was added but proved insufficient since the hardware capacity of the single instance was the limiting factor, illustrating a common scalability issue in distributed systems~\cite{tanenbaum2016distributed}. To address this, the second solution introduced database sharding~\cite{ozsu2011distributeddb}, which retained the core structure of the first design while distributing data across shards, thereby removing the database capacity constraint and supporting higher loads. However, new bottlenecks appeared in the machine learning service: the combined computational demand of inference and explanation generation for XAI overwhelmed the ML nodes, leading to high latency, request timeouts, and resource exhaustion. This showed that database scaling alone was insufficient, and the architecture required an asynchronous communication pattern to decouple ML processing from the main application flow.  

\begin{figure}
\centerline{\includegraphics[width=\columnwidth]{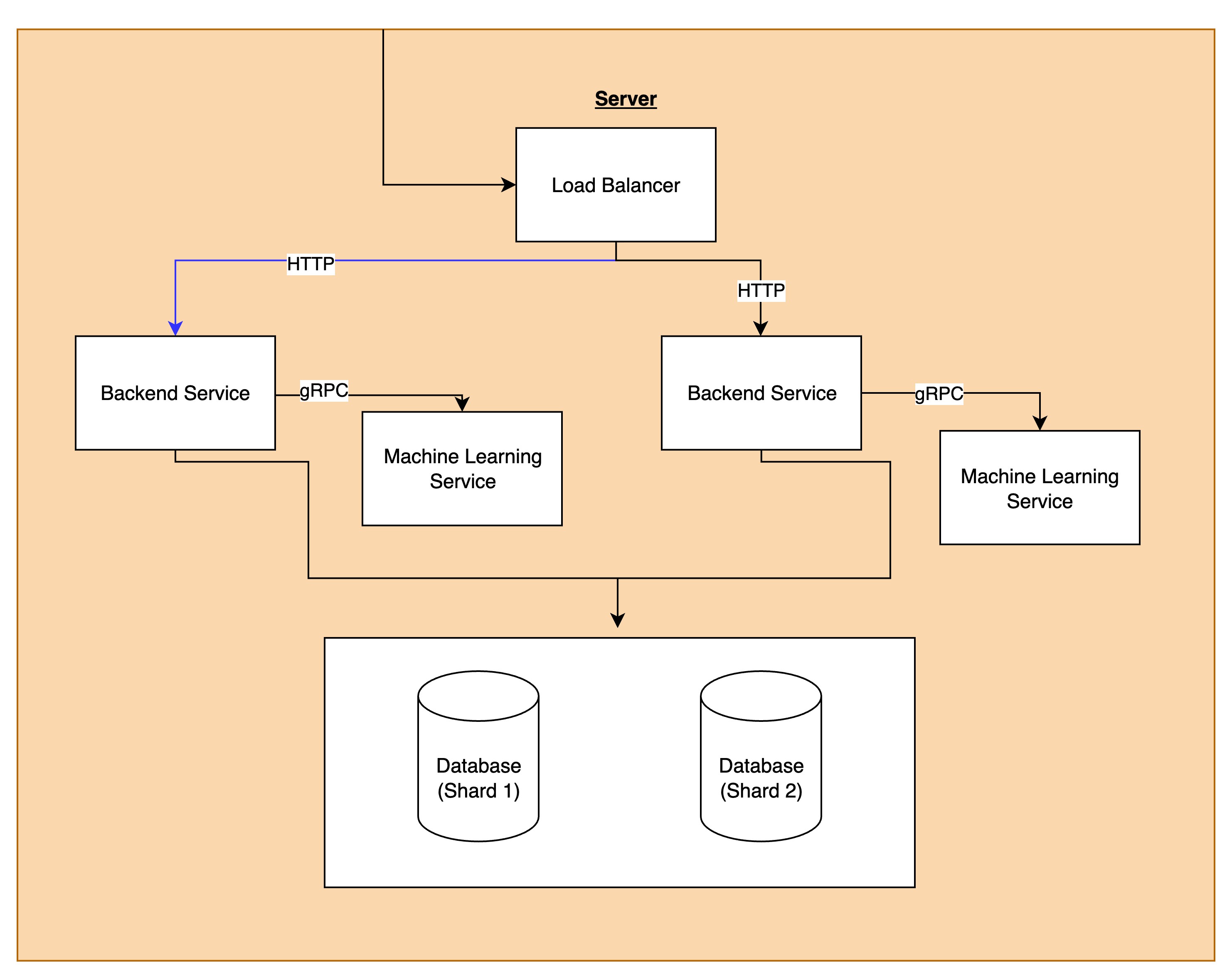}}
\caption{Architecture of Solution 2}
\label{fig_arch}
\end{figure}

\subsection{Final System Architecture}
\begin{figure}[h]
\centerline{\includegraphics[width=\columnwidth]{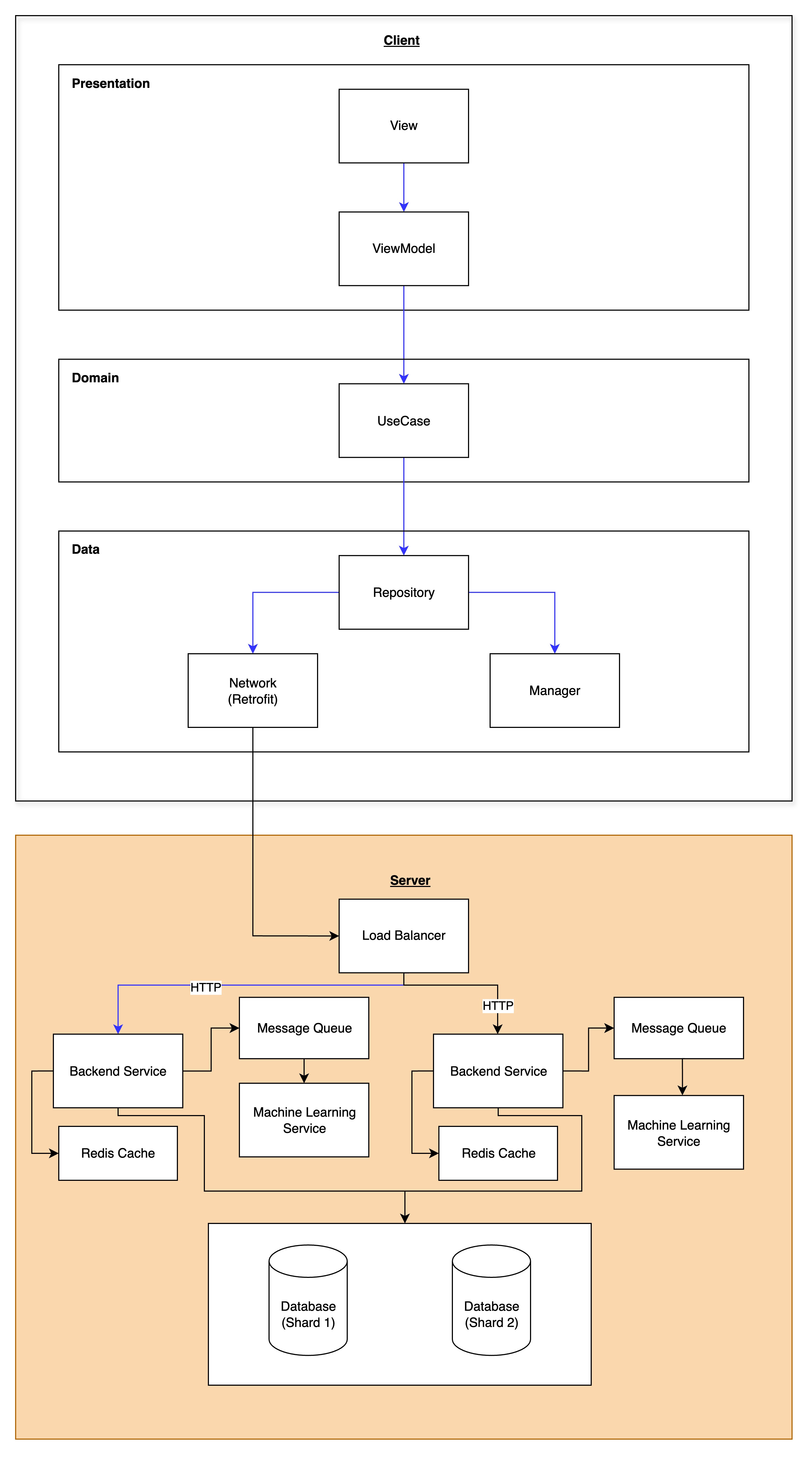}}
\caption{Final System Architecture Overview.}
\label{fig_arch}
\end{figure}

The final architecture, shown in Fig. \ref{fig_arch}, addresses the previously identified challenges by incorporating the chosen technology stack in a horizontally scaled design.

The server-side infrastructure is deployed across seven Virtual Private Servers (VPS), each with a 4-Core CPU and 8 GB of RAM with Ubuntu 20 operating system. This setup ensures distributed processing, fault tolerance, and improved performance under high load conditions. The key components and their interactions are:
\begin{enumerate}
    \item Load Balancer (1 VPS): Nginx distributes incoming client requests across back-end services. 
    \item Back-end Service (2 VPS): Stateless Go services handle the core application logic. For simplicity, Redis and RabbitMQ are deployed on the same VPS as the Go application. This node handles all requests related to user data manipulation, authentication, and other information, excluding prediction tasks.
    \item Machine Learning Service (2 VPS): Python services dedicated to AI model inference. This service focuses on handling prediction requests from the message queue.
    \item Database (2 VPS): PostgreSQL uses a range-based sharding strategy, partitioning user data across two shards by \texttt{user\_id}, with each shard handling 5,000 users. Each instance supports up to 1,000 connections, managed by PGBouncer to reduce overhead from concurrent transactions under heavy load.  
    \item Message Queue (RabbitMQ): For asynchronous prediction, the back-end acts as a \textit{producer}, publishing requests with a unique \texttt{CorrelationID} to the \texttt{ml.prediction.request} queue. The ML service, acting as a \textit{consumer}, processes the requests and sends results to the \texttt{ml.prediction.hybrid\_response} queue with the same \texttt{CorrelationID} for matching. All queues are configured as \texttt{durable} to ensure reliability.  
    \item Cache (Redis): Redis serves as an in-memory cache for temporary “what-if” prediction results, enabling fast simulations without repeatedly querying the main database. Cached data expires after one hour to maintain a balance between performance and freshness.  
\end{enumerate}

\subsection{Database Schema Design}
\begin{figure}[h]
\centerline{\includegraphics[width=\columnwidth]{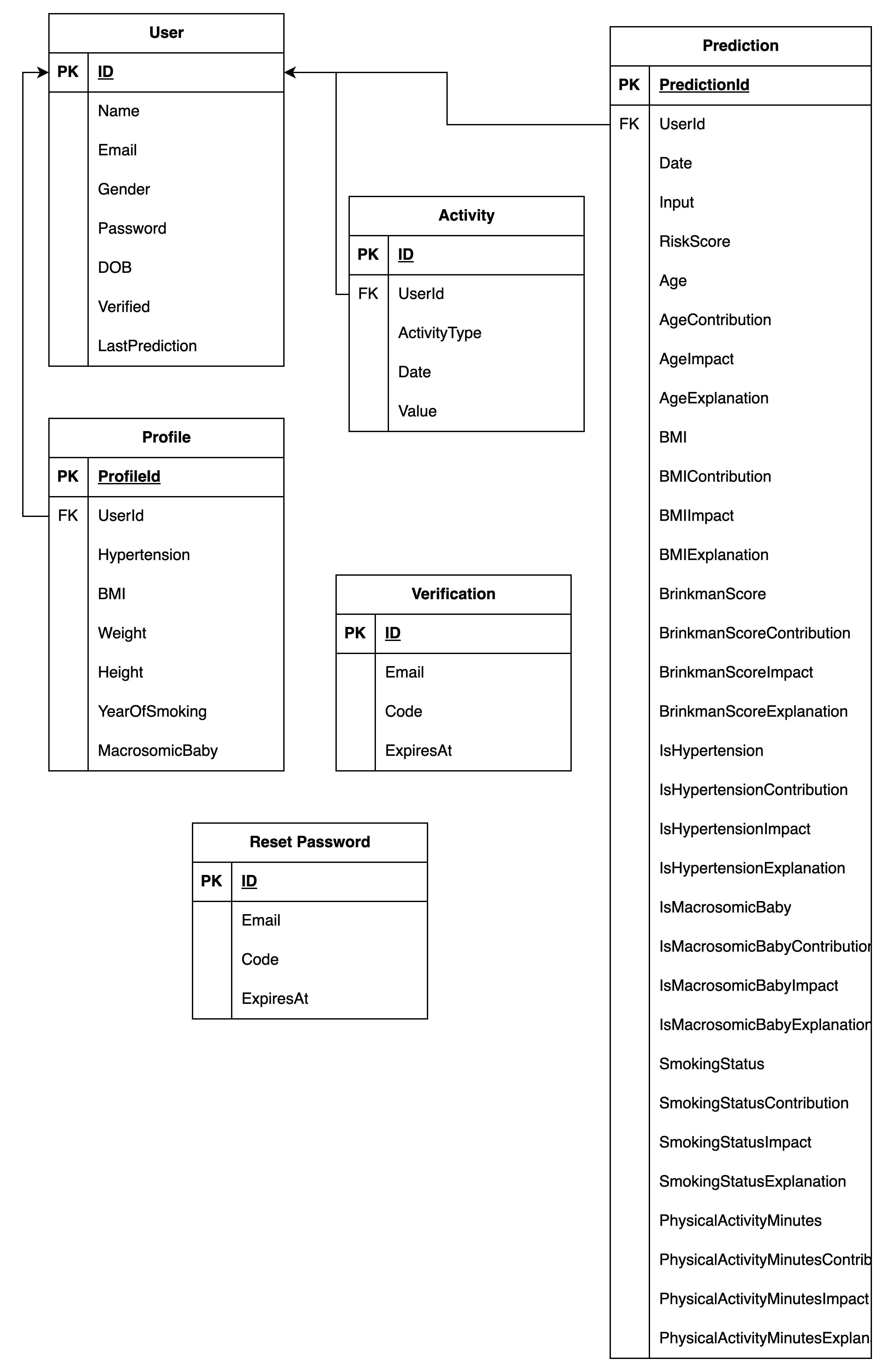}}
\caption{Entity Relationship Diagram (ERD) of the Application.}
\label{fig_erd}
\end{figure}

The database schema was designed to efficiently store all information related to users, their medical profiles, activities, verification, reset password, and prediction results. It follows a normalized structure to minimize redundancy and ensure data integrity across related entities. The Entity-Relationship Diagram (ERD) is shown in Fig. \ref{fig_erd}.

The most complex table is \texttt{Prediction}, which was intentionally designed with 33 attributes. This denormalized approach serves a specific purpose: to store not only the final risk score but also all the supporting data and descriptive text required for the XAI feature. Each row contains the prediction result along with the contribution, impact, and a natural language explanation for each factor (e.g., age, BMI, smoking status). This design allows the complete, explained prediction to be retrieved in a single, efficient query, which is crucial for a responsive user experience.

\section{Performance Evaluation}
The system's performance was rigorously evaluated to validate its ability to meet the predefined goals. The evaluation was conducted using the K6 load testing tool to simulate increasing user loads and measure key performance indicators (KPIs).

\subsection{Methodology}
The evaluation consisted of two parts:
\begin{enumerate}
    \item Performance Testing: Each of the 24 main API endpoints was subjected to an incremental load test, starting from 1,000 virtual users and increasing until performance degraded significantly. The primary metrics measured were average response time (latency), requests per second (throughput), and failure rate.
    \item Functional Testing: Unit tests were written for every controller function using the Testify library in Go to ensure business logic correctness in an isolated environment.
\end{enumerate}
The success criteria were defined as sustaining a load determined through scalability testing, with the system required to maintain an average latency of $\leq 1000$ ms and a failure rate of $\leq 5\%$ under peak conditions. Performance evaluation established that this threshold corresponded to approximately 10,000 concurrent virtual users.

\subsection{Results and Discussion}
The overall performance results at the 10,000-user mark are summarized in Table \ref{tab_perf}. Of the 24 features tested, 20 (83\%) achieved a ``PASS'' status, meeting both latency and failure rate targets. Four features received a ``PARTIAL'' status, meeting one of the two targets.

\begin{table}[t]
    \caption{Performance Evaluation Summary at 10,000 Users}
    \label{tab_perf}
    \begin{tabularx}{\columnwidth}{|>{\RaggedRight}X|c|c|c|}
        \hline
        \textbf{Feature (Endpoint)} & \textbf{Err. Rate (\%)} & \textbf{Avg. Latency (ms)} & \textbf{Status} \\
        \hline
        Login & 1.45\% & 1,100 & Partial \\
        Get User Info & 0.39\% & 601 & PASS \\
        Put User & 4.70\% & 691 & PASS \\
        Patch User & 2.76\% & 756 & PASS \\
        Get User Profile & 0.24\% & 1,100 & Partial \\
        Post User Profile & 0.56\% & 234 & PASS \\
        Put User Profile & 0.89\% & 167 & PASS \\
        Delete User Profile & 0.50\% & 243 & PASS \\
        Get Activity & 0.35\% & 56 & PASS \\
        Get Activity by ID & 0.94\% & 150 & PASS \\
        Post Activity & 0.13\% & 55 & PASS \\
        Put Activity & 1.26\% & 514 & PASS \\
        Delete Activity & 1.12\% & 717 & PASS \\
        Get Activity Count & 0.56\% & 69 & PASS \\
        Get Prediction & 1.54\% & 813 & PASS \\
        Get Prediction by Date & 3.53\% & 1,070 & Partial \\
        Post Prediction (Async) & 3.14\% & 1,539 & Partial \\
        Get Prediction Status & 1.27\% & 525 & PASS \\
        Get Prediction Result & 2.71\% & 787 & PASS \\
        Delete Prediction & 4.86\% & 945 & PASS \\
        \hline
    \end{tabularx}
\end{table}

\begin{enumerate}
    \item Core features, including \texttt{Get Activity} (56 ms) and \texttt{Post Activity} (55 ms), showed excellent performance with latencies well below 1\% failure rates. Throughput scaled linearly with load, demonstrating the effectiveness of Go, sharded PostgreSQL, and PGBouncer for CRUD operations.
    \item The \texttt{Post Prediction} endpoint revealed the need for asynchronous processing. A synchronous design failed under load, with over 31\% failures at 8,000 users and unstable response times, as back-end services were blocked by the overloaded ML service.
\end{enumerate}

\begin{figure}
    \centering
    \includegraphics[width=0.9\columnwidth]{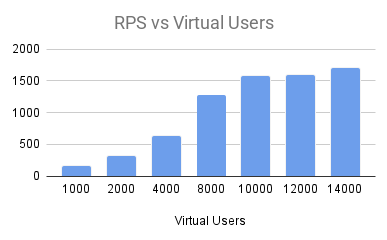}
    \caption{RPS vs Virtual Users for Get Activity Endpoint.}
    \label{fig_get_activity}
\end{figure}

\begin{figure}[h]
\centerline{\includegraphics[width=0.75\columnwidth]{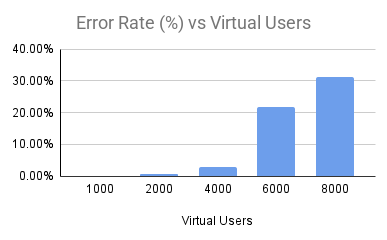}}
\caption{Error Rate of Synchronous Post Prediction.}
\label{fig_sync_error}
\end{figure}

\begin{figure}[h]
\centerline{\includegraphics[width=0.75\columnwidth]{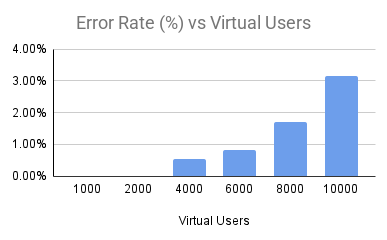}}
\caption{Error Rate of Asynchronous Post Prediction.}
\label{fig_async_error}
\end{figure}

Although the average latency for full asynchronous job completion was 1,539 ms above the target, this metric is somewhat misleading, as the initial API response acknowledging job submission was nearly instantaneous. By offloading long-running tasks, the system avoided server overload and ensured that no requests were dropped. Users could poll the \texttt{Get Job Status} endpoint, which had an average latency of 525 ms, to track progress. This trade-off between immediate results and system stability minimized error rates, improved reliability, and highlights a key principle of scalable architecture.

There are some risks to consider, especially with using a single load balancer, which could become a single point of failure. Additionally, encryption should be implemented for asynchronous communication between the backend service, machine learning components, and the message queue. The absence of rate limiting could also overwhelm the server if there’s a sudden spike in requests.

\section{Conclusion}
This research developed a scalable back-end architecture for an AI-powered diabetes prediction system. The platform achieved high reliability and performance, maintaining a sub-5\% failure rate and sub-1000 ms latency under 10,000 concurrent users.

\begin{enumerate}
    \item A horizontally scaled, microservices architecture ensures resilience and high throughput for both web and AI workloads.
    \item Asynchronous communication using RabbitMQ effectively manages long-running inference tasks and enhances fault tolerance.
    \item Database scalability through PostgreSQL sharding and PGBouncer reduces I/O bottlenecks in data-heavy operations.
    \item XAI can be integrated efficiently via decoupled, asynchronous workflows, maintaining transparency without reducing performance.
\end{enumerate}

Future improvements should explore adaptive load balancing, real-time monitoring, and stronger security measures such as end-to-end encryption and rate limiting. Overall, the architecture provides a robust foundation for large-scale health-tech systems.

\end{document}